%% file: cvpr.tex
\documentclass[10pt,twocolumn,letterpaper]{article}

\usepackage{CVPR2024/cvpr}              
\input{CVPR2024/preamble}

\usepackage{xcolor}
\definecolor{cvprblue}{rgb}{0.21,0.49,0.74}
\usepackage{hyperref}
\hypersetup{pagebackref,breaklinks,colorlinks,citecolor=cvprblue}

\usepackage{graphicx}
\usepackage[utf8]{inputenc} 
\usepackage[T1]{fontenc}    
\usepackage{hyperref}       
\usepackage{url}            
\usepackage{booktabs}       
\usepackage{amsfonts}       
\usepackage{amsmath}
\usepackage{amssymb}
\usepackage{nicefrac}       
\usepackage{microtype}      
\usepackage{subcaption}
\usepackage{wrapfig}
\usepackage{bbm} 
\usepackage{multirow}

\title{Convolutional Prompting meets Language Models for Continual Learning\vspace{-5mm}}

\author{Anurag Roy$^{1}$ \ \ \ Riddhiman Moulick$^{1}$ \ \ \ Vinay K. Verma$^{2}$\thanks{Work started before joining Amazon} \ \ \ Saptarshi Ghosh$^{1}$ \ \ \ Abir Das$^{1}$\\
$^{1}$IIT Kharagpur, $^{2}$IML Amazon India\\
\small{\texttt{\{anurag\_roy@,riddhimanmoulick@kgpian.,saptarshi@cse.,abir@cse.\}iitkgp.ac.in}}, \small{\texttt{vinayugc@gmail.com}}}

\begin{document}

\maketitle

\begin{abstract}
Continual Learning (CL) enables machine learning models to learn from continuously shifting new training data in absence of data from old tasks.
Recently, pretrained vision transformers combined with prompt tuning have shown promise for overcoming catastrophic forgetting in CL.
These approaches rely on a pool of learnable prompts which can be inefficient in sharing knowledge across tasks leading to inferior performance.
In addition, the lack of fine-grained layer specific prompts does not allow these to fully express the strength of the prompts for CL.
We address these limitations by proposing \model{}, a novel convolutional prompt creation mechanism that maintains layer-wise shared embeddings, enabling both layer-specific learning and better concept transfer across tasks.
The intelligent use of convolution enables us to maintain a low parameter overhead without compromising performance.
We further leverage Large Language Models to generate fine-grained text descriptions of each category which are used to get task similarity and dynamically decide the number of prompts to be learned.
Extensive experiments demonstrate the superiority of \model{} and improves SOTA by $\sim3\%$ with significantly less parameter overhead.
We also perform strong ablation over various modules to disentangle the importance of different components.
\footnote{Project page: \url{https://cvir.github.io/projects/convprompt}
}

\end{abstract}

\input{CVPR2024/intro}

\input{CVPR2024/related}

\input{CVPR2024/method}

\input{CVPR2024/exp}

\input{CVPR2024/ablation}

\input{CVPR2024/conclu}

\newpage
{
\small
\bibliographystyle{ieeenat_fullname}
\bibliography{cvpr}
}

\section*{Appendix}
\appendix

The appendix contains the following.
\vspace{1mm}
\begin{itemize}
	\setlength{\itemsep}{1mm}
	\item Section~\ref{sec:imp_details}: Finding the right Lambda value.
	\item Section~\ref{sec:image_sim_details}: More details about the Image Based  similarity.
	\item Section~\ref{sec:more_res}: Results of different tasks with and without task-similarity.
	\item Section~\ref{sec:similarity_exp}: A closer look into class attribute based task similarity.
\end{itemize}

\input{CVPR2024_supp/hyperparameter_search}

\input{CVPR2024_supp/image_sim_details}

\input{CVPR2024_supp/more_res}

\input{CVPR2024_supp/similarity_results}


\end{document}

%% file: CVPR2024/preamble.tex
\usepackage{color}
\definecolor{forest-green}{rgb}{0.59, 0.82, 0.47}
\definecolor{grey}{rgb}{0.93, 0.93, 0.93}
\definecolor{new-blue}{rgb}{0.5, 0.5, 0.99}
\definecolor{peach}{rgb}{0.99, 0.89, }
\definecolor{darkorchid}{rgb}{0.6, 0.2, 0.8}

\usepackage{soul}

\newcommand{\model}{\texttt{ConvPrompt}}

%% file: CVPR2024/intro.tex
\section{Introduction}
In this constantly changing world, computer vision models must adapt to the new and emerging concepts.
However, such models often suffer from catastrophic forgetting~\cite{plm1989catastrophic, goodfellow2015empirical, Robins1995CatastrophicFR}, a phenomenon where previously learned concepts are forgotten when adapting to novel concepts.
A trivial solution to this problem is to have separate models for each new task.
However, this would require task identities to be available during inference which may not be very practical.
Another way to tackle this will be to keep all the data that the model was trained on, add the new data to it, and then train the model again from the beginning.
Naturally, this will incur increased storage and computation as well.
A line of work attempting to maintain a balance, keeps a few samples from previous tasks and uses them for {\it rehearsing} the previous concepts while learning a new task~\cite{cvpr2017icarl,cvpr2019hou, cvpr2021der, lopez2017gradient, icpr2020ert,cvpr2021rm,hayes2020remind,prabhu2020gdumb,chaudhry2019continual,neurips2019rolnick}.
However, storing samples from older tasks may not always
be feasible, especially where long-term storage of
data is not permitted possibly due to privacy, security or legislative concerns~\cite{he2022exemplar}.
Therefore, developing {\it rehearsal-free} CL approaches requiring no storage of old data has come up to be desirable.

Recently, there has been a surge of prompt tuning based approaches~\cite{cvpr2022l2p, eccv2022dualprompt, cvpr2023coda, arxiv2023pop, iccv2023blurry, arxiv2023promptfusion, arxiv2023cpe, iccv2023langguidance, iccv2023prompt} that leverage on pre-trained transformers and show promising performance without using any rehearsal data.
Prompt tuning, originally introduced in NLP~\cite{liu2023pre} attaches small learnable vectors to a pre-trained frozen model to properly reuse the already learned representations.
Although promising for continual learning, these approaches, however, suffer from the following drawbacks -- (1)~Learning task-specific and task-shared information in separate layers ignores possible interaction between task-specific and task-shared components~\cite{verma2021efficient, iccv2023contracon} and (2)~Always learning a fixed number of prompts per task irrespective of the tasks' similarity with the previous ones.
The redundant prompts can lead to overfitting specifically in cases where the tasks are highly similar usually seen in fine-grained datasets.

In this work, we propose \model{}, which leverages task-specific prompts that is generated by convolution over task-shared parameters.
We also exploit similarity between tasks to control the generation of prompts.
Specifically, task shared knowledge is modeled by a learnable embedding matrix at each layer.
Prompts, on the other hand, are task specific and are generated by convoluting on the task shared embeddings with learnable kernels.
The shared embeddings are free to adapt with the different tasks.
The prompt generating convolution kernels, on the other hand, are set aside once learned for a task and new set of kernels are employed for the next task.
Such a design enables the shared embeddings to capture \textit{common concepts} while allowing the convolution operations to capture the \textit{task-specific concepts} from the common concepts.
Moreover, we employ language models to get similarity between tasks in order to determine the number of trainable convolution kernels.
Such an expansion strategy allows the model to be parameter efficient, introducing only the necessary number of new parameters as and when needed.
The benefits of our approach are three-fold -- (1)~It facilitates knowledge transfer across tasks using task-shared embeddings.
(2)~Convolutional prompt generation promotes efficient adaptation to new tasks with low parameter overhead.
(3)~Exploiting similar tasks by large language models results in further reduction in parameters and superior performance.

Extensive experimentation across several rehearsal-free continual learning benchmarks shows that our \model{} approach achieves significant performance gain compared to state-of-the-art approaches.
On average, across the 3 benchmark datasets and different experimental settings, we outperform the state-of-the-art prompt based CL approaches (\textit{e.g.}, CODA-Prompt~\cite{cvpr2023coda}) by a margin of $\sim 3\%$ while requiring a significantly lower number of parameters.
Our contribution can be summarized as follows:
\begin{itemize}
    \item We propose a local prompt creation mechanism by applying convolution over task-invariant global parameters enabling an efficient transfer of local and global concepts across tasks that helps the new tasks to adapt better.
    \item We incorporate a novel language based task similarity prediction, for the first time in continual learning, which helps to reduce the model parameters significantly without sacrificing the performance and without adding significant pre-processing overhead as well.
    \item The extensive ablations and experimentation over the various standard datasets and experimental settings show the superiority of our approach by a significant margin.   
\end{itemize}

%% file: CVPR2024/related.tex
\section{Related Work}

\noindent{\bf Continual Learning:} CL approaches can be classified into 3 broad categories.
(1)~\textit{Regularization-based methods} tackle catastrophic forgetting by applying regularizers that prioritize the preservation of important parameters associated with previously learned tasks.
By minimizing interference between new and old knowledge through the introduction of penalty terms~\cite{PNAS2017kirk, zenke2017continual, neurips2017lee, icml2017continual, eccv2018mas, cvpr2019task}  or constraining the direction of parameter update~\cite{iclr2021gpm, farajtabar2020orthogonal}, these methods encourage important parameters to remain in close proximity to previous solutions. While regularization based methods have shown promising results involving smaller number of tasks, their performance can be less satisfactory when confronted with challenging scenarios involving large number of tasks.
(2)~\textit{Dynamic Architecture-based methods} learn new tasks by assigning distinct parameters for each task~\cite{iclr2018den, cvpr2021der, cvpr2021_eft,cvpr2022meat}.
While these approaches exhibit the capability to learn extended sequences of tasks, they may encounter substantial memory and computational overhead.
Also, most approaches under this category require the information of which task an image belongs to during inference which may be unrealistic.
(3)~\textit{Rehearsal-based methods}~\cite{neurips2019er, douillard2021dytox, wang2022lvt, arxiv2023pop, cvpr2021rm, neurips2019gradient, aaai2018selective, icml2023birt} store a few representative training samples from previous tasks in a buffer, which is then used for training alongside the current task.
Though effective, these approaches are limited by the size of the buffer and the length of the task sequences.
These are also not particularly suitable for scenarios with data privacy requirements.
In contrast, \model{} addresses rehearsal-free continual learning by intelligently utilizing prompts on pre-trained models.

Recently, vision transformers have performed very well in CL~\cite{cvpr2022meat,douillard2021dytox,wang2022lvt, icml2023birt, iccv2023prompt, iccv2023contracon}.
Authors in~\cite{mirzadeh2022architecture} examined the impact of attention heads while Dytox~\cite{douillard2021dytox} acquires new skills by expanding special task tokens.
LVT~\cite{wang2022lvt} introduced inter-task attention in vision transformers.
Both methods, however, require additional memory to store previous training instances.
MEAT~\cite{cvpr2022meat} uses a parameter isolation approach while learning new tasks.
However, the model's expandability is limited, restricting the number of tasks it can learn and it requires task-ids to be known during inference.
ContraCon~\cite{iccv2023contracon} uses convolutional re-weighting of the self-attention weights.
However, inference is costly as it uses augmentation-based entropic task identification.

\noindent{\bf Prompt Learning}: Prompt-based CL offers robust protection against catastrophic forgetting by incorporating a small number of model instructions called prompts, rather than directly modifying encoder parameters~\cite{iccv2023prompt, cvpr2022l2p,cvpr2023coda,eccv2022dualprompt, arxiv2023pop, arxiv2023promptfusion, iccv2023langguidance, iccv2023blurry, arxiv2023cpe}.
Initial approaches like  L2P~\cite{cvpr2022l2p} and DualPrompt~\cite{eccv2022dualprompt} employ a prompt pool from which prompts are selected.
These methods match input data to prompts without relying on task identification, using local clustering based optimization.
Recently, S-Prompts~\cite{neurips2022sprompt} used prompts to continually learn in a domain incremental learning scenario, which involves learning the same set of classes under covariate distribution shifts.
CODA-Prompt builds on~\cite{cvpr2022l2p, eccv2022dualprompt} and applies soft attention on the prompts towards end-to-end continual learning. ProgressivePrompts~\cite{iclr2023progressiveprompts} progressively learns new prompt tokens for each incoming tasks, but assumes the presence of task-id during inference.
A contemporary work, LGCL~\cite{iccv2023langguidance}, uses handcrafted prompts from outside and contrastively learns to bring the output representation of the transformer to the prompt.
However, without indigenous prompt learning, this approach can act only as a plugin to existing CL approaches with incremental improvement in performance.
Our approach, \model{}, stands out from the rest in its proficiency in knowledge-sharing across tasks, its ability for on-the-fly prompt generation and its effective handling of the additional parameters required.

%% file: CVPR2024/method.tex
\input{CVPR2024/figs/convprompt_main_fig}
\section{Preliminaries}
\vspace{-1mm}
\label{sec:prelim}
\noindent{\textbf{Continual Learning}}:
Continual Learning (CL) trains models on tasks arriving sequentially over time without forgetting the previous ones.
Each task $t \in \{1, \ldots , T\}$ contains training samples $\{(x^t_i, y^t_i)\}$ where $x^t_i$ is the $i^{th}$ sample of the $t^{th}$ task and $y^t_i \in C^t$ is the corresponding label.
The set of the class labels for different tasks are mutually exclusive, \textit{i.e.}, $C^0 \cap C^1 \ldots \cap C^T = \phi$.
We address the challenging \textit{rehearsal-free} and \textit{Class-Incremental Learning} (CIL) setting of CL where a  trained model $f$ needs to predict the label $y = f(x)$ for an unseen test sample $x$, regardless of its task and without access to training data from previous tasks.

\noindent{\textbf{Transformer Architecture}}:
We build our approach on top of a pre-trained Vision Transformer (ViT)~\cite{iclr2021vit}.
A transformer starts by dividing an input image $x$ into a set of $N$ fixed-sized patches $z_1 \in \mathbb{R}^{N \times d}$ which are then embedded into a $d$-dimensional space with positional encoding.
A single encoder layer of ViT consists of stacks of Multi-Head Self-Attention (MHSA), layer normalization and Feed Forward Network (FFN) blocks with residual connections.
Given the input $z_l$ at the $l^{th}$ layer, the output $z_{l+1}$ is generated that goes into the next layer as input for $l \in \{1, 2, \cdots L\}$ where $L$ is the total number of encoder layers.
At the $l^{th}$ layer, MHSA block computes self-attention on the input $z_l$ by using $H$ separate self-attention heads.
Self-attention values from head $h \in \{1,2, \cdots H\}$ at layer $l$ is given by,
\vspace{-3mm}
\begin{align}
    \label{eq:sa}
    A(Q_{l,h}, K_{l,h}, V_{l,h}) & = softmax \bigg(\frac{Q_{l,h} K^T_{l,h}}{\sqrt{d_k}}\bigg) V_{l,h}
    \vspace{-3mm}
\end{align}
where $Q_{l,h}=z_l W^Q_{l,h}$,  $K_{l,h} = z_l W^K_{l,h}$ and $V_{l,h} = z_l W^V_{l,h}$ are query, key, and value with learnable weights $W^Q_{l,h}, W^K_{l,h}$ and $W^V_{l,h} \in \mathbb{R}^{d\times d_h}$ respectively.
$d_h = d/H$ is the dimension of key, query and value vectors.
The activations from different attention heads are then concatenated and residually added to the input $z_l$ before performing layer normalization.
The resulting activations are passed through a FFN block consisting of two linear layers and an activation function (usually GELU).
After another residual connection and layer normalization, the output $z_{l+1}$ at the $l^{th}$ layer is generated.

\noindent{\textbf{Prompt and  Prefix Tuning}}:
Prefix or prompt tuning aims to learn continuous vectors where the pre-trained transformer is kept frozen.
It prepends $l_p$ learnable vectors to the original keys and values of the self-attention heads at every layer.
Specifically, $l_p$ length prefix vectors $P^{(K)}_{l,h};P^{(V)}_{l,h} \in \mathbb{R}^{l_p\times d_h}$ are concatenated with the original key $K_{l,h}$ and value $V_{l,h}$ respectively.
Then the self-attention values from head $h$ at layer $l$ is computed as $A(Q_{l,h}, [P^{(K)}_{l,h}, K_{l,h}], [P^{(V)}_{l,h}, V_{l,h}])$ following Eqn.~\ref{eq:sa}.
Unlike existing prompt based approaches learning directly the additional vectors~\cite{cvpr2022l2p,eccv2022dualprompt,cvpr2023coda}, our work focuses on creating prompts by maintaining a balance between old knowledge of the system and new information to be put on an ad-hoc basis.

\vspace{-2mm}
\section{Methodology}
\label{sec:methodology}

In this work, we propose a prompt-based CL approach (\model{}) by generating prompt vectors for each new task in combination with knowledge learned previously (ref. Fig.~\ref{fig:prompt_cl_fig2}).
Knowledge from previous tasks is modeled by a \textit{learnable embedding matrix shared between all tasks}.
The task-specific prompts are created by employing convolution operation on the shared embedding matrix.
While existing works~\cite{cvpr2022l2p,eccv2022dualprompt} have shown to perform well by prompting a pre-trained transformer, they rely on a single set of prompt vectors needing to compress all necessary information of an image into one single set.
Rather than one set of $l_p$ prompt vectors, following CODA-Prompt~\cite{cvpr2023coda}, we have $M$ such sets which we call \emph{prompt components}.
However, unlike CODA-Prompt, the prompt components are \textit{not} directly learned in our approach.
Instead, they are generated from previous task knowledge by employing $M$ convolution kernels  (known as the \emph{prompt generators}) in each head of each layer.
A weighted combination of the prompt components provides the final $l_p$ prompt vectors where the weights come from the \texttt{[CLS]} embedding at each layer of the ViT corresponding to the input image.
The weighing not only allows us to optimize the model end-to-end but also provides a unique blend of previous task knowledge and the input image.

\subsection{Prompt Generation}
\label{sec:pc}

The prompt vectors in our approach are dynamically generated by small convolution operations.
Convolution is performed between two learnable components -- (i) Shared Embeddings and (ii) Prompt Generators.
Corresponding to each head $h \in \{1,2, \cdots H\}$ in each layer $l \in \{1,2, \cdots L\}$ of the transformer, there are shared embedding matrices $SE^{K}_{l,h}$ and $SE^{V}_{l,h}$ respectively for the keys and the values.
These matrices are shared across tasks.
The prompt generators are single channel convolution kernels which are applied on the shared embeddings.
We learn a set of $M$ prompt generators for each head in each layer, for both keys and values.
For head $h$ and layer $l$, the prompt generators are denoted as $G^{K}_{l,h,m}$ and $G^{V}_{l,h,m}$, where $m \in \{1,2, \cdots M\}$.
The shared embeddings are of dimension $(l_p+k-1)\times (d_h+k-1)$ where $k$ is the size of the convolution kernels of the prompt generators.
The convolution operation results in prompt components $PC^K_{l,h,m}$ and $PC^V_{l,h,m}$ of size $l_p\times d_h$.
The convolutional prompt generators not only enables us to maintain a good trade-off between performance and low parameter requirements per task but also makes use of the inherent inductive bias for structured prompt creation~\cite{tsai2023convolutional}.

\subsection{Prompt Weighting}
\label{sec:pcw}
Instead of compressing all the information into one set of prompt vectors we make use of $M$ prompt components to get the final prompt vectors at each head of each layer.
For each prompt component $PC^K_{l,h,m}$ (and $PC^V_{l,h,m}$), we generate weights (between $-1$ and $1$) which can be interpreted as the relative importance of the particular prompt component in blending them together.
We employ $M$ learnable keys, referred to as \emph{prompt keys} $\pi \in \mathbb{R}^{d_\pi}$ for this purpose.
The prompt keys work on the image features expressed as a nonlinear function of the \texttt{[CLS]} token at each layer.
The \texttt{[CLS]} token is passed through a Projection Network ($PN_{\phi}$), parameterized by $\phi$, consisting of a two-layer fully-connected neural network with a ReLU activation~\cite{icml2010relu} in between.
The output $PN_{\phi}(\texttt{[CLS]})$ maps the input image to the same space as the prompt keys where a cosine similarity is taken between $PN_{\phi}(\texttt{[CLS]})$ and the $M$ prompt keys in each layer $l$ resulting in the importance values $\{s_{l,1}, s_{l,2}, \cdots, s_{l,M}\}$.
The prompt components obtained as a result of the convolution operation between the prompt generators and shared embeddings are weighed by the $M$ similarity scores to get the final prompts $P^{(K)}_{l+1,h}$ and $P^{(V)}_{l+1,h}$ as follows,
\begin{equation}
\label{eq:promt_final_k}
    P^{(K)}_{l+1,h} \!\!=\!\! \sum_{m=1}^{M} \!\!s_{l,m} PC^K_{l,h,m}; P^{(V)}_{l+1,h} \!\!=\!\! \sum_{m=1}^{M} \!\!s_{l,m} PC^V_{l,h,m}
\end{equation}

While existing works have linearly combined prompt components to get the final prompt, the similarity scores were generated using the same final \texttt{[CLS]} token across all layers.
However, this requires a full forward pass through the transformer solely for getting the similarity scores and then the final prompts at each layer for the final predictions resulting in a total of two passes.
In contrast, we utilize the \texttt{[cls]} embedding from each layer enabling us to generate the final prompts for the subsequent layers in one single pass resulting in a significant reduction of computation.
The similarity values generated from the image help different prompt components to focus on specific features towards the final prompt~\cite{cvpr2023coda}.
Building on this, we used a non-linearly learned projection network that better captures complex tasks as shown empirically.

\input{CVPR2024/tables/results_tab_cifar_cub}

\subsection{Language Guided Prompting}
To maintain a balance between learning new tasks and preserving knowledge accumulated from old tasks, we make use of both task-shared as well as task-specific parameters.
In our work, $PN_{\phi}$ and $SE$ act as shared parameters facilitating inter-task information sharing, while with new tasks coming, the previously learnt prompt generators and prompt-keys are frozen, thus making them task-specialized.
A new set of these are freshly learned for every incoming task.

Let the total number of prompt generators per layer learnt till the $(t-1)^{th}$ task be $M_{t-1}$, while the number of prompt generators learnt for task $i$ only is $J_i$.
Naturally, $M_{t-1} = \sum\limits_{i=1}^{t-1} J_i$ and all $M_{t-1}$ prompt generators are kept frozen when the new set of $J_t$ prompt generators are learned.
Notwithstanding the prompt generators' role to mitigate catastrophic forgetting, it is also crucial to keep the increase in parameters in check with increasing number of tasks.
Ideally, if a task is similar to a task seen earlier, then the prompt generators learnt previously can be reused.
Thus, in contrast to previous works~\cite{cvpr2023coda} which learn a fixed number of prompt components per task, we learn a dynamic number of prompt generators depending on the similarity of the task with the previous ones.
The task similarity can be naively modeled by comparing the visual features of the images from different tasks.
We present an alternative framework to get task similarity cheaply using Large Language Models (LLMs) such as GPT-3~\cite{brown2020language} which show remarkable world knowledge on a variety of topics.

Our key insight is that we can use \textit{language} as a tool to get descriptions of visual attributes of different classes and use these to find task similarity.
Visual attributes are additional semantic knowledge that articulate the visual concepts associated with each categories.
For example, some visual attributes of \emph{bee} are `black and yellow stripes', `two pairs of wings', `three body segments', \textit{etc}.
Instead of manually writing these, we queried GPT-3 to get the attributes for each set of classes in each task as they arrived.
This is computationally cheap, requires no additional training, and is scalable to large number of classes.
Inspired by works like~\cite{arxiv2022visual,cvpr2023doubly}, the class attributes are generated by using the query -- ``{\it What are useful features for distinguishing a [class name] in a photo?}'' for each class in a task.
We generate the BERT embeddings of these attributes and store them in a pool for all seen tasks.
For each attribute of the current task $t$, cosine similarity with all stored attribute embeddings till task $t-1$ is computed.
The similarity $sim_t$ of task $t$ with the previous tasks, is the mean of such maximum similarity values across all attributes of task $t$.
Let $J_{max}$ denote the maximum prompt generators per task.
Then $J_t$ is given by $(1-sim_t)J_{max}$ \textit{i.e.}, higher similarity enables to have lower number of prompts.
Leveraging linguistic knowledge, we reduce the learnable parameters if the classes in a task have a high overlap with the previously encountered ones.

\subsection{Regularization and Final Objective}
To prevent over-writing of concepts captured by previous tasks in global task-shared $PN_{\phi}$ and $SE$, we need to ensure that while learning the current task, these parameters deviate less from the previous tasks. To achieve this,  when learning for task $t$, we regularize the set of parameters $\phi_t$ of the projection network and the shared embeddings $SE_t$, to have low $l_1$ norm with that of the previous task as follows:
\begin{equation}
\label{eq:regu}
\begin{split}
        \mathcal{L}_r(\phi_t, \phi_{t-1}) &= ||\phi_{t-1} - \phi_{t}||_1 \\
        \mathcal{L}_r(SE_t, SE_{t-1}) &= ||SE_{t-1} - SE_{t}||_1
\end{split}
\end{equation}
$SE_t$ denote the shared embedding parameters of task $t$ for all heads and layers combined and for both keys and values.
Note that suffix $t$ used with the shared embeddings may indicate that $SE$ is task specific, they are not.
\emph{Shared} signifies the same set of embeddings is used to learn the \emph{shared} semantics among different tasks.
We incrementally update the same SE for each task, using a copy of the SE from the immediately preceding task for regularization in Eqn.~\ref{eq:regu} that is discarded after training.
$\phi_t$ denote the projection network parameters at the $t^{th}$ task.
The final objective is:
\small
\begin{equation}
  \label{eq:loss_fn}
  \mathcal{L}_{cls}(f(x), y)\!+\!\mathbbm{1}\!(t\!>\!1)\lambda [\mathcal{L}_r(\phi_{t},\!\phi_{t-1})\!+\!\mathcal{L}_r(SE_{t},\!SE_{t-1})]
\end{equation}
\normalsize
where $\mathcal{L}_{cls}$ denotes classification loss, $t$ the task-id and $\lambda \in [0, 1]$ the hyper-parameter to weigh the loss components.
The indicator function $\mathbbm{1}\!(t\!>\!1)$ denotes the fact that regularization is applied after the first task.

%% file: CVPR2024/figs/convprompt_main_fig.tex
\begin{figure*}[t!]
    \centering
    \includegraphics[scale=0.25]{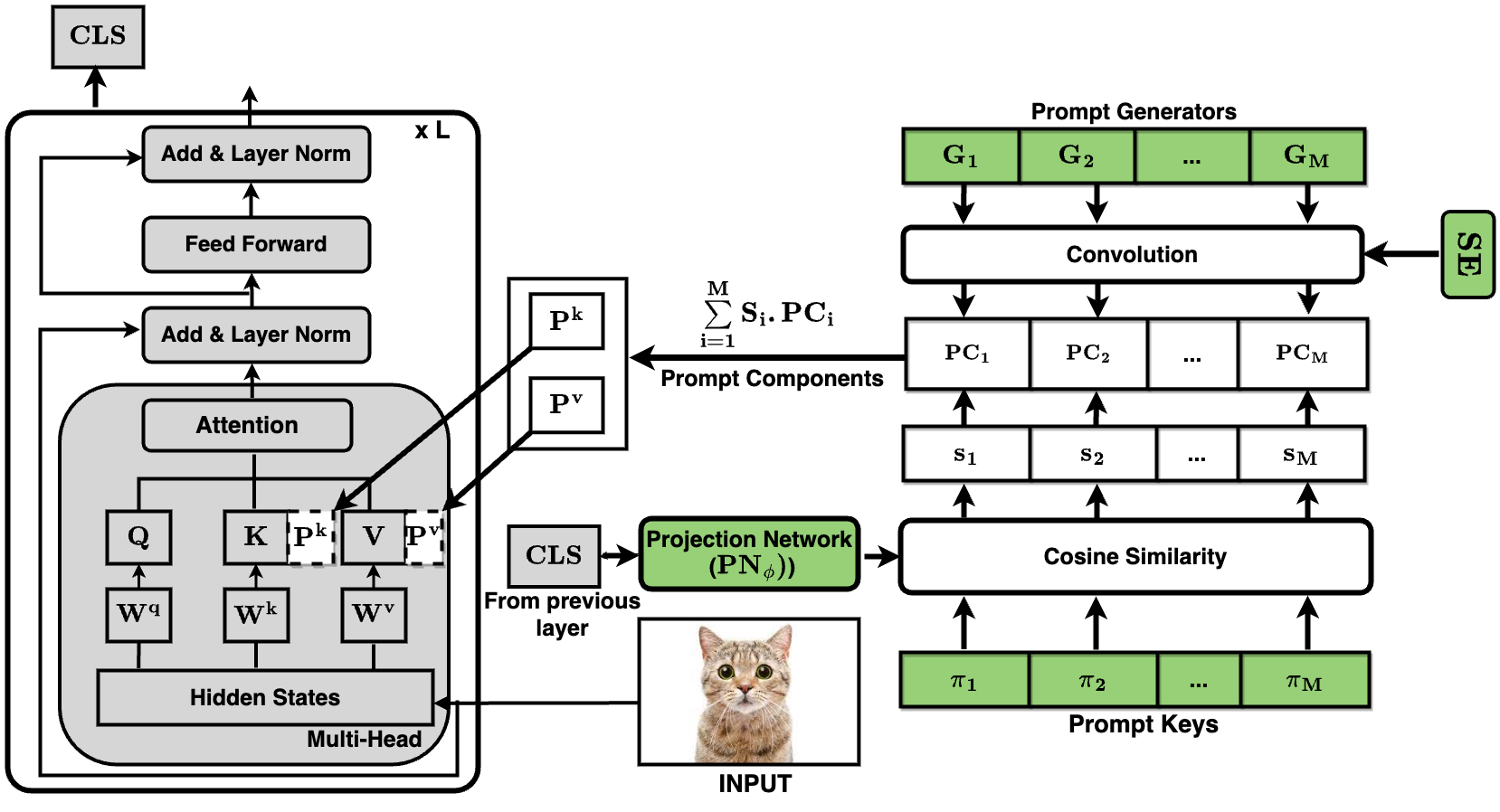}
    \vspace{-2mm}
    \caption{\small {\bf The proposed \model{} architecture.} The layerwise \texttt{[cls]} embedding is passed through the projection network to generate task-specific query representations which are then matched with the prompt-keys using cosine similarity to get the similarity values of prompt-generators. The prompt generators are applied over the shared embeddings to generate input specific prompt components. A weighted average of these  prompt components are calculated with the corresponding cosine similarity value as weights to get the input specific key and value prompts which are applied to the pre-trained model. The learnable components are highlighted in {\bf green}, the frozen components are highlighted in {\bf grey}.}
    \label{fig:prompt_cl_fig2}
    \vspace{-5mm}
\end{figure*}

%% file: CVPR2024/tables/results_tab_cifar_cub.tex
\begin{table*}[t!]
    \centering
    \small
     \addtolength{\tabcolsep}{8.0pt}
    \begin{tabular}{l||ll||ll||l}
    \hline
    {\bf Tasks} & \multicolumn{2}{c}{\bf Split CIFAR-100} &  \multicolumn{2}{c}{\bf Split CUB-200} & {\bf $N_{param} (\downarrow)$}\\
    \hline
    {\bf Method} & {\bf $A_T ( \uparrow )$} & {\bf $F_T ( \downarrow )$} & {\bf $A_T ( \uparrow )$} & {\bf $F_T ( \downarrow )$} & {\bf Train/Total}\\
    \hline
    Joint-FT (upper bound) & $93.22 \pm 0.16$ & \textendash & $88.00 \pm 0.15$ & \textendash & $100 /\ 100$\\
    \hline

    Seq-FT & $8.6 \pm 0.43$ & $42.67 \pm 0.13$ & $23.87 \pm 0.54$  & $62.52 \pm 0.57$ & $100/\ 100$ \\
    ER (buffer size 5000) & $82.30 \pm 0.42$ & $16.30 \pm 0.24$ & $60.73 \pm 0.23$ & $8.71 \pm 0.65$ & $100/\ 100$ \\
    LwF~\cite{tpami2017lwf} & $64.56 \pm 1.23$ & $25.27 \pm 1.32$ & $48.73 \pm 1.46$ & $25.18 \pm 0.31$ & $100/\ 100$ \\
    
    L2P~\cite{cvpr2022l2p} & $82.76 \pm 1.17$ & $7.86 \pm  0.39$ & $62.21 \pm 1.92$ & $7.12 \pm 0.33$ & $\mathbf{0.7/\ 100.7}$ \\
    L2P + LGCL ~\cite{iccv2023langguidance} & $84.33 \pm 0.06$ & $5.83 \pm  0.23$ & \textendash & \textendash &$\mathbf{0.7/\ 100.7}$ \\
    DualPrompt~\cite{eccv2022dualprompt} & $85.07 \pm 0.49$ & $5.57 \pm 0.20$ & $66.00 \pm 0.57$ & $\mathbf{4.4 \pm 0.31}$ & $1.3/\ 101.3$ \\ 
    DualPrompt + LGCL~\cite{iccv2023langguidance} & $87.23 \pm 0.21$ & $5.10 \pm 0.15$ & \textendash & \textendash & $1.3/\ 101.3$ \\ 
    
    CODA-Prompt~\cite{cvpr2023coda} & $87.00 \pm 0.38$ & $4.78 \pm 0.24$ & $74.40 \pm 0.74$ & $6.40 \pm 0.34$ & $4.6/\ 104.6$ \\
    \hline
    \model{} & $\mathbf{88.87 \pm 0.33}$ & $\mathbf{4.75 \pm 0.15}$ & $\mathbf{80.2 \pm 0.52}$ & $\underline{5.6 \pm 0.38}$ & $2.0 /\ 102.0$ \\
    \hline
    \end{tabular}
    \caption{{\bf Results (\%) on CIFAR-100 and CUB-200}. Reported results are for $10$ tasks with a supervised ImageNet-21k pretrained ViT as the backbone. $A_T$ denotes the average accuracy and $F_T$ denotes the forgetting. $N_{param}$ denotes the percentage of trainable/final parameters w.r.t that of the ViT model. The ${N_{param}}$ values are dynamic for our approach and the reported value ($2.0/102.0$) is the average of the values in Split-CIFAR-100 ($2.2/102.2$) and Split CUB-200 ($1.8/101.8$).} 
    \label{tab:results_cifar_cub}
    \vspace{-5mm}
\end{table*}

%% file: CVPR2024/exp.tex
\section{Experiments}
\label{sec:exp}

\input{CVPR2024/tables/results_tab_imr}

\noindent{\textbf{Datasets}}:
We evaluate \model{} on the benchmark datasets, ImageNet-R~\cite{iccv2021imagenetr} and CIFAR-100~\cite{2009cifar}, as well as on the fine-grained dataset, CUB-200~\cite{WahCUB_200_2011} in the CIL setup.
ImageNet-R~\cite{iccv2021imagenetr} is formed from 200 subcategories of ImageNet~\cite{russakovsky2015imagenet} but with images from different domains such as cartoon, graffiti and origami. 
It also includes some hard examples from ImageNet that standard models fail to classify.
It contains 24,000 training images and 6,000 test images. 
Following~\cite{iccv2023slca, eccv2022dualprompt}, we split the $200$ classes into $10$ tasks with each task containing $20$ classes respectively.
CIFAR-100~\cite{2009cifar}, a widely used dataset in continual learning, contains $100$ classes, with each having $500$ training and $100$ test images.
Following~\cite{iccv2023slca,cvpr2022l2p,eccv2022dualprompt, cvpr2023coda}, we use the $10$ task setup of CIFAR-100 with each task containing $10$ classes. 
CUB-200~\cite{WahCUB_200_2011} is a fine-grained dataset containing $200$ classes of different bird species with $5994$ training images and $5794$ test images.
Following~\cite{iccv2023slca}, we use the $10$ task setup of CUB-200 with each task containing $20$ classes.

\vspace{1mm}
\noindent{\textbf{Training and Implementation Details}}:
We use the ViT-B/16~\cite{iclr2021vit} model pre-trained on the ImageNet-21k~\cite{ridnik2021imagenet21k} as the backbone over which \model{} is applied.
Our projector network is a two-layer neural network having $d/2$ and $d/4$ neurons in these two layers respectively where the input (\texttt{[CLS]} token) is $d$-dimensional.
A ReLU~\cite{icml2010relu} activation function is applied between the two layers.
Our approach applies prompts to $7$ layers of the pre-trained ViT as our ablation experiments show that more layers with prompts does not help improve the performance although parameter overhead increases.
We train each task in CIFAR-100, ImageNet-R and CUB-200 for $10$, $10$ and $60$ epochs respectively. 
The hyperparameter $J_{max}$ is set to 5, acting as an upper limit for the maximum number of prompt components needed per task.
The hyperparameter $\lambda$,  which is used to weigh the regularization terms in Eqn.~\ref{eq:loss_fn} is set to $0.01$.
The experiments leading to our choice of $\lambda$ and other additional experimental results can be found in the supplementary material.
We present our results after conducting five random trials, where task orders were randomly selected for each run. The $mean \pm std$ values are reported.

\vspace{1mm}
\noindent{\bf Metrics Used:} We report Average accuracy $A_T$ and Forgetting $F_T$ calculated over the $T$ tasks for all our experiments. Specifically, after training on $T$ tasks is completed, $A_T$ and $F_T$ are calculated as follows:
\begin{equation}
\begin{split}
    A_T &= \frac{1}{T} \sum_{t=1}^{T} S_{t,T} \\
    F_T &= \frac{1}{T-1} \sum_{t=1}^{T-1} \underset{t' \in \{1, \ldots, T-1\}}{max} (S_{t,t'} - S_{t,T})
\end{split}
\end{equation}
where $S_{t,T}$ is the test classification accruacy on task $t$ after the model has been trained on task $T$.
In other words, average accuracy measures the average accuracy of all the tasks after training on the last task and forgetting measures the average drop in accuracy of a task after training on the last task from its maximum accuracy attained.

\vspace{1mm}
\noindent{\bf Comparisons:} We evaluated our approach against several rehearsal free approaches.
These include Learning without Forgetting (LwF)~\cite{tpami2017lwf}, Learning to Prompt (L2P)~\cite{cvpr2022l2p}, DualPrompt~\cite{eccv2022dualprompt} and CODA-Prompt~\cite{cvpr2023coda}.
We also evaluated against LGCL~\cite{iccv2023langguidance}, which uses language based prompts and acts as a plugin to L2P and DualPrompt.
Additionally, we also compared our approach with a rehearsal-based approach, Experience Replay (ER)~\cite{arxiv2019er} with buffer size $5000$.
In addition, we report Joint-FT and Seq-FT performances as they serve as bounds of the performance in many situations.
In Joint-FT, the ViT model is trained jointly on the training data of all the tasks combined, serving as an upper bound on the performance.
Seq-FT represents fine-tuning the ViT model sequentially using only the new task's training data and thus is severely affected by catastrophic forgetting.

\subsection{Results and Analysis}
As shown in Table~\ref{tab:results_cifar_cub} and Table~\ref{tab:results_imr}, \model{} outperforms both rehearsal-free and rehearsal-based approaches significantly.
On average, \model{} outperforms the existing state-of-the-art CODA-Prompt~\cite{cvpr2023coda} by $\sim3\%$ while using only $\sim40\%$ of the trainable parameters used by CODA-Prompt.
Our approach shows slightly more forgetting in the $10$-task setup of Split-CUB.
However, even with more forgetting, our approach is able to outperform the existing approaches by atleast $3\%$ margin thereby confirming that our design enables efficient adaptation to new tasks by preventing overfitting that leads to higher maximum accuracy by the tasks.
This indicates that our convolutional prompt creation mechanism is able to utilize the shared inter-task concepts better than CODA-Prompt~\cite{cvpr2023coda} and DualPrompt~\cite{eccv2022dualprompt}. 

\input{CVPR2024/tables/slca_tab}

\noindent{\textbf{Comparison with SLCA}}:
We also conducted a comparison with a recent SOTA approach \emph{Slow Learner with Classifier Alignment} (SLCA)~\cite{iccv2023slca} that does not use prompts for CL.
Instead, it finetunes the whole network with smaller learning rate for the represenation layers and larger learning rate for the classification layer.
This approach has demonstrated superior performance compared to existing continual prompt-tuning methods.
As it involves full network tuning, it is computationally expensive making it impractical in resource-constrained scenarios.
In contrast, our approach is viable in situations where compute is limited.
In this experiment, we show that our approach can also exploit differential learning rates and classifier alignment as SLCA and provide further improvement in performance.
Specifically, in such cases, we learn the prompts along with the transformer weights during fine-tuning.
The results in Table~\ref{tab:slca} demonstrate that our approach combined with SLCA, outperforms it in two out of the three datasets achieving the new state-of-the-art while in the CIFAR-100 dataset it is almost at par with SLCA.

\input{CVPR2024/tables/ablation_tab}

%% file: CVPR2024/tables/results_tab_imr.tex
\begin{table*}[t!]
    \centering
    \small
     \addtolength{\tabcolsep}{0.5pt}
     \resizebox{\textwidth}{!}{
    \begin{tabular}{l||ll||ll||ll||l}
    \hline
    {\bf Tasks} & \multicolumn{2}{c}{\bf 5 Task} & \multicolumn{2}{c}{\bf 10 Task} & \multicolumn{2}{c}{\bf 20 Task} & {\bf $N_{param} (\downarrow)$}\\
    \hline
    {\bf Method} & {\bf $A_T ( \uparrow )$} & {\bf $F_T ( \downarrow )$} & {\bf $A_T ( \uparrow )$} & {\bf $F_T ( \downarrow )$} & {\bf $A_T ( \uparrow )$} & {\bf $F_T ( \downarrow )$} & {\bf Train/Total}\\
    \hline
    Joint-FT (upper bound) & $79.6 \pm 0.87$ & \textendash & $79.6 \pm 0.87$ & \textendash & $79.6 \pm 0.87$ & \textendash & $100 /\ 100$\\
    \hline
    Seq-FT & $21.82 \pm 0.85$ & $ 76.26 \pm 0.37 $ & $11.42 \pm 0.76$ & $78.32 \pm 0.64$ & $8.75 \pm 0.42$ & $82.21 \pm 0.96$ & $100/\ 100$ \\
    ER (buffer size 5000) & $70.53 \pm 0.68$ & $17.47 \pm 0.35$ & $64.32 \pm 0.65$ & $22.35 \pm 0.97$ & $53.26 \pm 0.83$ & $34.21 \pm 0.82$ & $100/\ 100$ \\
    LwF~\cite{tpami2017lwf} & $49.75 \pm 0.65$ & $41.36 \pm 0.27$ & $39.27 \pm 1.92$ & $51.23 \pm 0.34$ & $30.29 \pm  1.82$ & $60.32 \pm 0.86$ & $100/\ 100$ \\
    L2P~\cite{cvpr2022l2p} & $67.43 \pm 0.11$ & $5.12 \pm 0.62$ & $63.49 \pm 0.40$ & $6.85 \pm 0.42$ & $59.38 \pm 0.50$ & $5.89 \pm 0.36$ & $\mathbf{0.7/\ 100.7}$ \\
    L2P + LGCL ~\cite{iccv2023langguidance} & \textendash & \textendash & $62.51 \pm 0.05$ & $8.9 \pm  0.17$ & \textendash & \textendash &$\mathbf{0.7/\ 100.7}$ \\
    DualPrompt~\cite{eccv2022dualprompt} & $70.42 \pm 0.88$ & $4.1 \pm 0.33$ & $68.50 \pm 0.52$ & $5.14 \pm 0.18$ & $63.21 \pm 0.49$ & $5.28 \pm 0.45$ & $1.3/\ 101.3$ \\ 
    DualPrompt + LGCL~\cite{iccv2023langguidance} & \textendash & \textendash & $69.46 \pm 0.04$  & $4.2 \pm 0.06$ & \textendash & \textendash & $1.3/\ 101.3$ \\
    CODA-Prompt~\cite{cvpr2023coda} & $75.38 \pm 0.34$ & $6.08 \pm 0.36$ & $74.24 \pm 0.56$ & $4.92 \pm 0.21$ & $70.86 \pm 0.42$ & $6.87 \pm 0.25$ & $4.6/\ 104.6$ \\
    \hline
    \model{} & $\mathbf{79.10 \pm 0.47}$ & $\mathbf{3.08 \pm 0.11}$ & $\mathbf{77.86 \pm 0.25}$ & $\mathbf{4.33 \pm 0.24}$ & $\mathbf{75.1 \pm 0.39}$ &  $\mathbf{4.1 \pm 0.29}$ & $2.2 /\ 102.2$ \\
    \hline
    \end{tabular}}
    \caption{{\bf Results (\%) on ImageNet-R}. Reported results are for $5$, $10$ and $20$ tasks splits of ImageNet-R with a supervised ImageNet-21k pretrained ViT as the backbone. $A_T$ denotes the average accuracy and $F_T$ denotes the forgetting. $N_{param}$ denotes the percentage of trainable/final parameters w.r.t that of the ViT model. The ${N_{param}}$ values vary for \model{} with varying number of tasks and the reported value is the average of the values for 5-tasks ($1.64/101.64$), 10-tasks ($2.0/102.0$) and 20 tasks ($2.88/102.88$).}
    \label{tab:results_imr}
    \vspace{-3mm}
\end{table*}

%% file: CVPR2024/tables/slca_tab.tex
\begin{table}
    \centering
    \footnotesize
     \addtolength{\tabcolsep}{0.8mm}
    \begin{tabular}{l||l|l}
    \hline
    {\bf Dataset} & {\bf SLCA} & {\bf SLCA + \model{}}  \\
    \hline
    {\bf Split CIFAR-100} & $91.53 \pm 0.28$ & $90.60 \pm 0.35$ \\

    {\bf Split ImageNet-R} & $77.00 \pm 0.33$ & $78.5 \pm 0.37$ \\
    {\bf Split CUB-200}  & $84.71 \pm 0.40$ & $87.12 \pm 0.31$ \\
    \hline
    \end{tabular}
   \vspace{0mm}
    \caption{{\bf Results with SLCA:} We report the $A_T$ values for SLCA and the SLCA + \model{}. \model{} when applied on top of SLCA, improves its performance by $1-2\%$.}
    \vspace{-3mm}
\label{tab:slca}
\end{table}

%% file: CVPR2024/tables/ablation_tab.tex
\begin{table}
    \centering
    \footnotesize
     
     \resizebox{0.48\textwidth}{!}{
     \addtolength{\tabcolsep}{-3pt}
    \begin{tabular}{l||ll|l}
    \hline
    
    {\bf Method} & {\bf $A_T ( \uparrow )$} & {\bf $F_T (\downarrow) $}  & {\bf $N_{param} (\downarrow)$}\\
    \hline
    Upper Bound & $79.7 \pm 0.15$ & \textendash  & $100/100$\\ 
    \hline
    ViT (linear probing) (Lower Bound) & $62.1 \pm 0.52$ & $ 5.74 \pm 0.23$ & $0.02 / 100$ \\
    \hline
    + $SE$  & $67.3 \pm 0.58$ & $5.19 \pm 0.21$ & $0.7/100.7$\\
    + $SE$ + NN & $73.82 \pm 0.84$ & $9.92 \pm 0.48$ & $14.9/114.9$\\
    + $SE$ + Conv & $73.92 \pm 0.65$ & $9.43 \pm 0.33$ & $3.4/103.4$\\
    
    + $SE$ + Conv + $PN$ (1-layer) & $76.28 \pm 0.23$ & $4.46 \pm 0.17$ & $3.5/103.5$ \\
    + $SE$ + Conv + $PN$ (2-layer) & $77.96 \pm 0.54$ & $4.75 \pm 0.57$ & $3.7/103.7$ \\
    \hline
    ViT + $SE$ + Conv + & & & \\
    $PN$ (2-layer) + Task-Sim & $77.86 \pm 0.25$ & $4.33 \pm 0.24$ & $2.0/102.0$ \\
    ( \model{} ) & & & \\
    \hline 
    \end{tabular}}
   \vspace{0mm}
    \caption{{\bf Ablation over ImageNet-R 10 tasks:} We report $A_T$ and $F_T$ averaged for $5$ trials. We also report the number of trainable params of each of the variants. ViT denotes the ViT pre-trained on ImageNet-21k over which we apply the different components.}
    \vspace{-3mm}
\label{tab:ablation}
\end{table}

%% file: CVPR2024/ablation.tex
\subsection{Ablation Studies and Other Analysis}
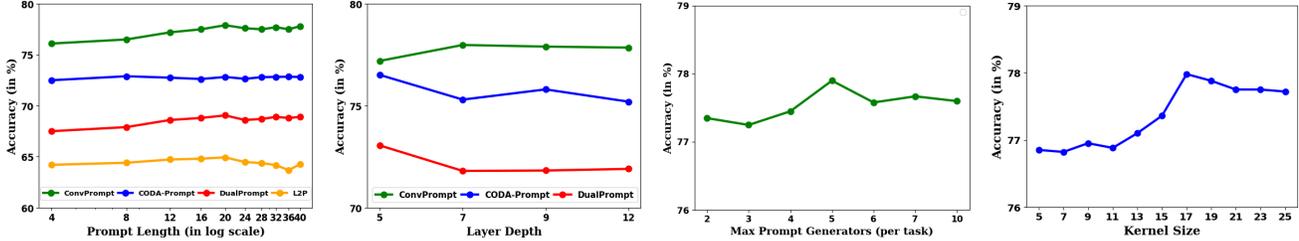
\begin{figure*}[!t]
\captionsetup[subfigure]{aboveskip=-1pt,belowskip=-1pt}
\input{CVPR2024/figs/prompt_len_ablation}
\input{CVPR2024/figs/prompt_depth_ablation}
\input{CVPR2024/figs/prompt_creators_ablation}
\input{CVPR2024/figs/kernel_size_ablation}
\vspace{-4mm}
\caption{{\bf Ablation over Split ImageNet-R 10 tasks:} (a)~average accuracy vs prompt length: The performance peaks at $8$ for CODA-P and at $20$ for rest of the models (b)~average accuracy vs number of layers prompts are applied to: The performance for \model{}, peaks at $7$ layers while it peaks at $5$ for the other models.
(c)~Average accuracy vs maximum number of prompt components per task: The performance of \model{} peaks at $J_{max}=5$ prompts per task, with the final prompt count after 10 tasks reaching to $18$ owing to task-similarity. (d) Average accuracy vs kernel size of the prompt creators: The performance for \model{}, peaks at  kernel size $17$.}
\vspace{-1\baselineskip}
\label{fig:ablation}
\end{figure*}

We perform all our ablation studies on the $10$ task setup of the  ImageNet-R dataset unless otherwise mentioned.
Throughout the study, we consider the ImageNet-21k pre-trained ViT-B/16 model onto which we gradually integrate our modules to showcase their importance.
Table~\ref{tab:ablation} shows the results.

\noindent{\textbf{Exploiting Shared Embedding across Tasks}}:
Adding shared embedding on top of the ViT-B/16 baseline, already gives better performance than L2P~\cite{cvpr2022l2p} that uses a pool of prompts (ref. row `$+SE$').
This shows that knowledge shared across tasks, when properly regularized, can itself be better than purely individual prompts.

\noindent{\textbf{Role of Prompt Generators}}:
Next, we add the task-specific prompt generators on top of the task-shared embeddings.
We experimented with two functional forms of the prompt generators -- a) Convolution kernels and b) Neural Networks.
Both types show significant improvement (ref. rows `$+SE+NN$' and `$+SE+$Conv').
However, using a neural network is naturally heavier on compute compared to a convolutional prompt generator as shown in the rightmost column.
Overall, the addition of task-specific prompt generators leads to a significant improvement of $\sim6\%$ over the task-shared-only approach and an improvement of $\sim3\%$ over DualPrompt~\cite{eccv2022dualprompt} which also uses both task shared and task specific knowledge.
This reaffirms our hypothesis that prompt generation over shared embeddings helps in the effective sharing of inter-task concepts to better adaptation of newer tasks.

\noindent{\textbf{Significance of Prompt Weighing}}:
We then add our prompt weighting mechanism which further gives an improvement of $\sim 3\%$ (ref. two rows starting with `+ $SE$ + Conv + $PN$').
This signifies that our prompt weighting exploits important prompts better.
To investigate if a non-linearity in projector network helps, we compare the performances of a linear projection network (single layer fully-connected neural net) with a non-linear one (two-layer fully-connected neural net with a ReLU in between) and observe that the non-linearity naturally is better with complex visual data albeit with a slight increase in parameter count.

\noindent{\textbf{Language helps reduce parameters}}:
Finally, we add language driven task-similarity to dynamically determine the number of prompt generators (ref. last row of Table~\ref{tab:ablation}) which makes the full \model{}.
This leads to significant parameter reduction while maintaining the same performance.
In datasets containing very similar tasks (\textit{e.g.}, CUB-200) the performance also gets boosted ($\sim 1\%$) possibly due to less overfitting as a result of reduced number of parameters.

\noindent{\textbf{Effect of Prompt Length}}:
We analyze the effect of the length ($l_p$) of prompt vectors on the performances of different prompt based CL approaches including ours.
For this purpose, we ran experiments with different prompt lengths, increasing the length in multiples of $4$ from $4$ to $40$.
As can be seen in Fig.~\ref{fig:ablation}\textcolor{red}{a}, \model{} performs the best across all values of $l_p$ and the performance has an increasing trend till the prompt length is 20 after which it saturates.
So, we used $l_p = 20$, unless otherwise mentioned.

\noindent{\textbf{Effect of Increasing Layers to Prompt}}:
To analyze if prompting every layer helps we applied prompts to different layers of the pre-trained backbone for \model{} and the closely related approaches (ref. Fig.~\ref{fig:ablation}\textcolor{red}{b}).
Specifically, starting with prompts to the first 5 layers, we go on to apply prompts till the last layer (\textit{i.e.}, $12^{th}$ layer).
As seen, \model{}'s performance peaks at $7$ layers before stagnating while for DualPrompt and CODA-Prompt performance decreases after $5$ layers and then stagnates.

\noindent{\textbf{Effect of number of Prompt Generators}}:
We analyze the effect of increasing the maximum number of prompt generators $J_{max}$ (ref. Fig.~\ref{fig:ablation}\textcolor{red}{c}).
We observe that the performance peaks at $J_{max} = 5$ after which the performance decreases.

\noindent{\textbf{Effect of Kernel Size of Prompt Generators}}:
To understand the impact of the prompt generator's kernel size, we varied the convolution kernel size $k$ from $5$ to $25$ in steps of $2$.
As can be seen in Fig.~\ref{fig:ablation}\textcolor{red}{d}, convolution kernels of size $17$ gives the best results while slightly decreasing and stagnating for higher values.
We chose kernel size $17$ as the default value since it leads to a better trade-off between the number of trainable parameters and performance.
\input{CVPR2024/tables/task_sim_tab}

\input{CVPR2024/tables/prefix_before_after}

\noindent{\textbf{Effect of Prompting before or after Projection}}: We analyze the effect of concatenating the prompt vectors $P^{(K)}_{l,h}$ and $P^{(V)}_{l,h}$ with the original key $K_{l,h}$ and value $V_{l,h}$ respectively vis-a-vis concatenating them with the input $z_l$ which is projected to get the key and values.
Performance does not vary much if projection is performed after concatenation ($C\!\!\rightarrow\!\!P$) compared to the other way round ($P\!\!\rightarrow\!\!C$) (Table~\ref{tab:results_pret}).
Naturally, computation is more (by $\sim 0.2B$ MACs) in ($C\!\!\rightarrow\!\!P$) due to bigger matrix-vector multiplications.
With low compute overhead and comparable performance, ($P\!\!\rightarrow\!\!C$) is advantageous and is used in our experiments.

\noindent{\textbf{Best way to measure Task Similarity}}:
As similarity between tasks plays a crucial role in performance as well as additional parameters trained, we tried different avenues to measure task similarity.
Specifically, we experimented with (i) class-label based task similarity and (ii) image-based task similarity.
In the first approach, instead of taking GPT-3 generated attributes, we directly used the class labels and in the second we extracted visual features using pretrained ViT-B/16 and used them to measure similarity with previous tasks.
Table~\ref{tab:task_similarity} shows the performance and parameter requirements with these on 10-task ImageNet-R.
It can be seen that the attribute-based class similarity leads to the introduction of least number of parameters as well as best performance.
\input{CVPR2024/tables/inference_times}

\noindent{\textbf{Comparing Inference-Time}}:
We compared the inference time with competing prompt tuning based approaches, namely L2P~\cite{cvpr2022l2p}, DualPrompt~\cite{eccv2022dualprompt} and CODA-Prompt~\cite{cvpr2023coda}. We compute the number of MACs (Multiply and Accumulate Operations) during inference after the model has been trained for all the $10$ tasks on the ImageNet-R dataset.
As is seen in Table~\ref{tab:inference_time}, \model{} requires the least compute backed by the single pass for prompting through it.

%% file: CVPR2024/figs/prompt_len_ablation.tex
\begin{subfigure}[b]{0.24\textwidth}
    \centering
    \includegraphics[width=\linewidth]{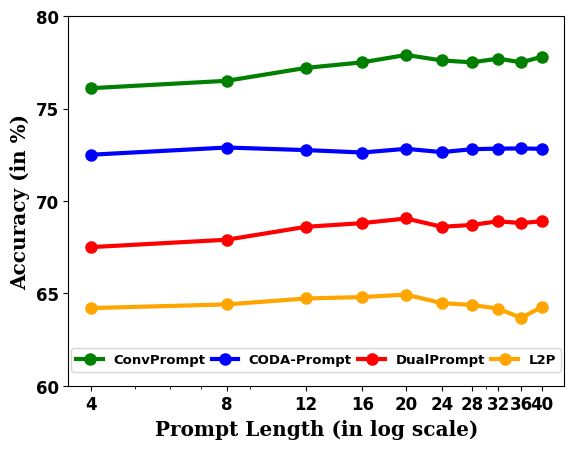}
    \label{fig:promptlen_ablation}
\end{subfigure}

%% file: CVPR2024/figs/prompt_depth_ablation.tex
\begin{subfigure}[b]{0.24\textwidth}
    \centering
    \includegraphics[width=\linewidth]{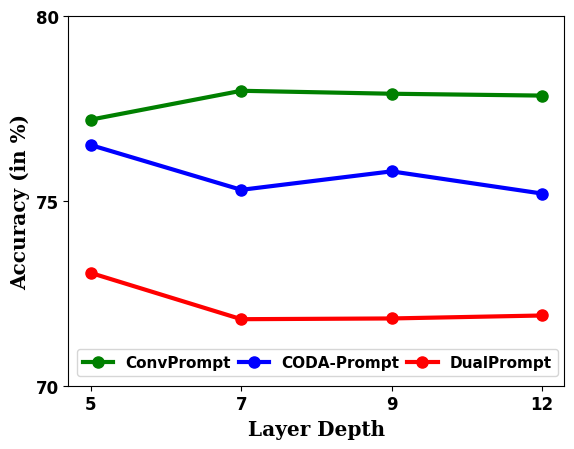}
    \label{fig:prompt_depth_ablation}
\end{subfigure}

%% file: CVPR2024/figs/prompt_creators_ablation.tex
\begin{subfigure}[b]{0.24\textwidth}
    \centering
    \includegraphics[width=\linewidth]{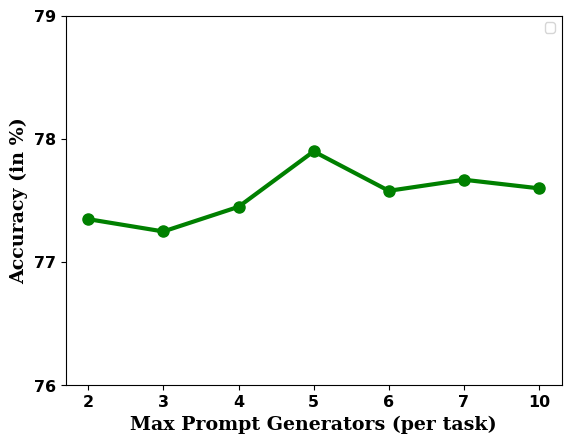}
    \label{fig:prompt_creator_ablation}
\end{subfigure}

%% file: CVPR2024/figs/kernel_size_ablation.tex
\begin{subfigure}[b]{0.24\textwidth}
    \centering
    \includegraphics[width=\linewidth]{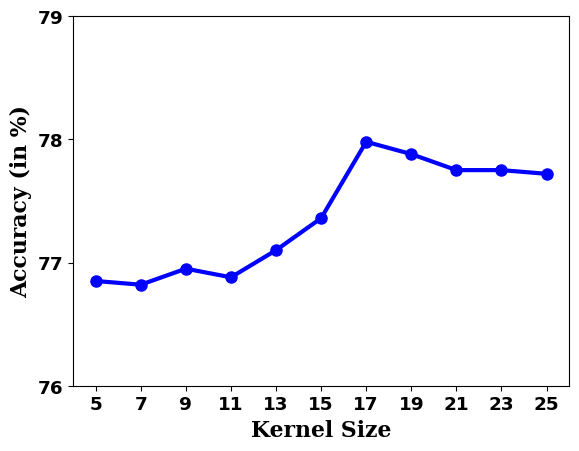}
    \label{fig:ker_size_ablation}
\end{subfigure}

%% file: CVPR2024/tables/task_sim_tab.tex
\begin{table}
    \centering
    \footnotesize
     \addtolength{\tabcolsep}{0.8mm}
     \resizebox{0.48\textwidth}{!}{
    \begin{tabular}{l||ll|l}
    \hline
    {\bf Method} & {\bf $A_T ( \uparrow )$} & {\bf $F_T (\downarrow) $} &{\bf $N_{param} (\downarrow)$}\\
    \hline
    class-label based sim & $76.93 \pm 0.45$ & $4.49 \pm 0.57$ & $2.22/102.22$ \\
    image-based sim & $77.18 \pm 0.42$ & $4.38 \pm 0.47$ & $2.29/102.29$ \\
    \hline
    Class Attribute  & & & \\
    based sim (\model{}) & $77.86 \pm 0.25$ & $4.33 \pm 0.24$ & $2.00/102.00$ \\
    \hline
    \end{tabular}
    }
   \vspace{0mm}
    \caption{{\bf  ImageNet-R 10 tasks:} Comparison of different task-similarity measures. We report $A_T$ and $F_T$ averaged for $5$ trials.}
    \vspace{-4mm}
\label{tab:task_similarity}
\end{table}

%% file: CVPR2024/tables/prefix_before_after.tex
\begin{table}[h!]
    \centering
    \scriptsize
     \addtolength{\tabcolsep}{-1.5pt}
     \vspace{-3mm}
     \resizebox{0.48\textwidth}{!}{
    \begin{tabular}{l||ll||ll}
    \hline
     & \multicolumn{2}{c||}{\bf $P\!\!\rightarrow\!\!C$} &  \multicolumn{2}{c}{\bf $C\!\!\rightarrow\!\!P$} \\
    \hline
    {\bf Datasets} & {\bf $A_T ( \uparrow )$} & {\bf $F_T ( \downarrow )$} & {\bf $A_T ( \uparrow )$} & {\bf $F_T ( \downarrow )$} \\
    \hline
    {\bf CIFAR-100} & $88.87 \pm 0.33$ & $4.75 \pm 0.15$ & $88.24 \pm 0.31$ & $3.86 \pm 0.34$ \\
    {\bf ImageNet-R} & $77.86 \pm 0.25$ & $4.33 \pm 0.24$ & $77.76 \pm 0.28$ & $3.65 \pm 0.27$ \\
    {\bf CUB-200} & $80.2 \pm 0.52$ & $5.6 \pm 0.38$ & $80.1 \pm 0.45$ & $5.7 \pm 0.26$ \\ \hline
    \end{tabular}}
    \vspace{-2mm}
    \caption{{Prefixes before and after projection on 10-task setup.}}
    \label{tab:results_pret}
    \vspace{-3mm}
\end{table}

%% file: CVPR2024/tables/inference_times.tex
\begin{table}
    \centering
    \footnotesize
     \addtolength{\tabcolsep}{10.5mm}
    \begin{tabular}{l||l}
    \hline   
    {\bf Method} & {\bf MACs}  \\
    \hline
    L2P~\cite{cvpr2022l2p} & $35.85B$ \\
    DualPrompt~\cite{eccv2022dualprompt} & $33.72B$ \\
    CODA-Prompt~\cite{cvpr2023coda}  & $33.72B$ \\
    \model{} & $17.98B$ \\
    \hline
    \end{tabular}
   \vspace{0mm}
    \caption{{\bf Comparison of Inference Times:} MAC (multiply accumulate) operations required for the evaluation of tasks once the model has been fully trained on the 10 tasks of  Split ImageNet-R.}
    \vspace{-3mm}
\label{tab:inference_time}
\end{table}

%% file: CVPR2024/conclu.tex
\section{Conclusion}
In this paper, we proposed \model{},  a novel convolutional prompt generation mechanism coupled with a task similarity based expansion strategy for rehearsal-free CL.
Different from the existing approaches, our approach creates prompts in each layer by applying convolution over task shared embeddings  causing better knowledge transfer across tasks.
Moreover, our expansion strategy with LLM driven task similarity ensures that this performance boost is achieved without a significant increase in the number of learnable parameters.
Extensive experimentation showed that \model{} outperforms SOTA baselines significantly while requiring fewer additional parameters. 


%% file: CVPR2024_supp/hyperparameter_search.tex
\section{Sensitivity Analysis of $\lambda$}
\label{sec:imp_details}
\begin{figure}[ht!]
    \centering
    \includegraphics[scale=0.5]{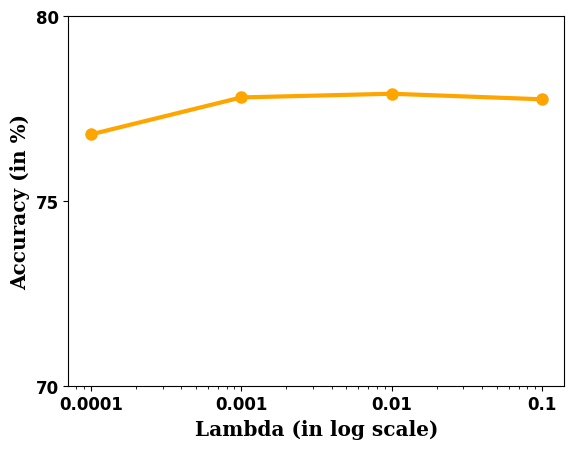}
    \caption{\small {\bf Average Accuracy ($\mathbf{A_T}$) vs. $\mathbf{\lambda}$ value(Log Scaled) plot for \model{}}. The performance peaks at $1.00E-02$ and saturates thereafter.}
    \label{fig:lambda_val}
\end{figure}

In this section, we discuss in more detail, about the hyper-parameter search, especially the $\lambda$ value for weighing the $\mathcal{L}_{norm}$ regulariser. As shown in Fig:~\ref{fig:lambda_val}, the performance of the model increases from $1.00E-4$ to $1.00E-3$ peaks at $1.00E-02$ and saturates around that value, hence the $\lambda$ value for \model{} is chosen as $0.01$.

%% file: CVPR2024_supp/image_sim_details.tex
\section{Image-based Similarity Calculation}
To calculate image-based task similarity, we fetch the image features of each of the images of the classes in the tasks seen by the model till now. We take class-wise average of these features and store them in a pool of seen classes. Therefore, for each class we have a single embedding representative for all images belonging to that class. For extraction of image features we use the final layer \texttt{[cls]} embeddings of the ViT-B/16 model, pre-trained on ImageNet-21K. In order to find the task-wise similarities, we compute the cosine similarity of the embeddings of all classes that had arrived in the previous tasks ($0$ to $t-1$) with the embeddings of the classes in task $t$. For each of the classes in task $t$, their maximum similarity score with any of the previous classes is considered. Finally, the similarity of task $t$ is computed as the maximum similarities averaged over all the classes in task $t$.
\label{sec:image_sim_details}

%% file: CVPR2024_supp/more_res.tex
\section{With and Without Task Similarity}
\input{CVPR2024_supp/figs/similarity_images}

We report the comparison on the performance  and the number of parameters required by the \model{} with and without language based task similarity based attribute reduction. The results with and without task similarity remain almost similar, while a ~50\% reduction in parameters is observed with task-similarity. Furthermore for datasets with very similar tasks, such as CUB-200~\cite{WahCUB_200_2011}, task-similarity also improves performance by ~1\%, by preventing overfitting.

\begin{table}[ht!]
    \centering
    \footnotesize
     \addtolength{\tabcolsep}{0.8mm}
    \resizebox{0.48\textwidth}{!}{
    \begin{tabular}{l||ll|ll}
    \hline

    \multirow{2}{*}{\bf Tasks} & \multicolumn{2}{c}{\bf Without Sim} & \multicolumn{2}{c}{\bf With Sim} \\
    \cline{2-5}
    & {\bf $A_T$} & {\bf $N_{params}$} & {\bf $A_T$} & {\bf $N_{params}$} \\
    \hline
    {\bf Split CIFAR-100} & $88.91 \pm 0.29$ & $3.72/103.72$ & $88.87 \pm 0.33$ & $2.2/102.2$\\
    {\bf Split ImageNet-R} & $77.96 \pm 0.54$ & $3.7/103.7$ & $77.86 \pm 0.25$ & $2.0/102.0$ \\
    {\bf Split CUB-200} & $79.25 \pm 0.38$ & $3.68/103.68$ & $80.2 \pm 0.52$ & $1.8/101.8$ \\
    
    \hline
    \end{tabular}}
   \vspace{0mm}
    \caption{{\bf Results with and without task-similarity:} We report the $A_T$ values for 10 task trials for each of the datasets averaged over 5 trials.
    }
\label{tab:slca}
\end{table}
\label{sec:more_res}

%% file: CVPR2024_supp/figs/similarity_images.tex
\begin{figure*}[ht!]
\captionsetup[subfigure]{aboveskip=-1pt,belowskip=-1pt}
\begin{subfigure}[b]{\textwidth}
	\centering
	\includegraphics[scale=0.16]{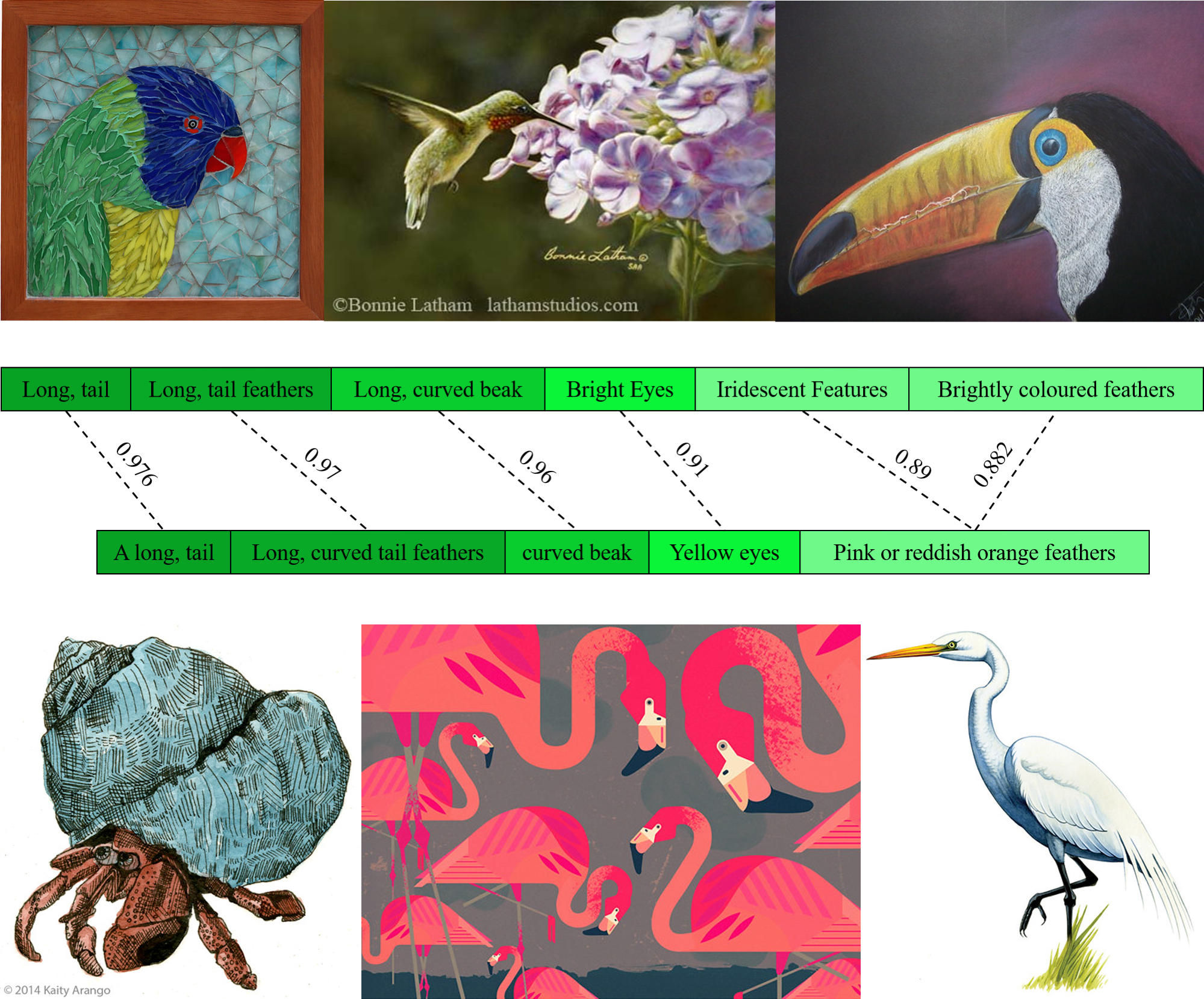}
	\vspace{1mm}
	\caption{{\bf An example of two similar tasks:} {\bf Task 1} contains the classes \emph{lorikeet}, \emph{hummingbird}, \emph{toucan} and {\bf Task 2} contains the classes \emph{hermit crab}, \emph{flamingo}, and \emph{american egret}.}
	\label{fig:sim2_figs}
	\vspace{2mm}
\end{subfigure}
\begin{subfigure}[t!]{\textwidth}
	\centering
	\includegraphics[scale=0.16]{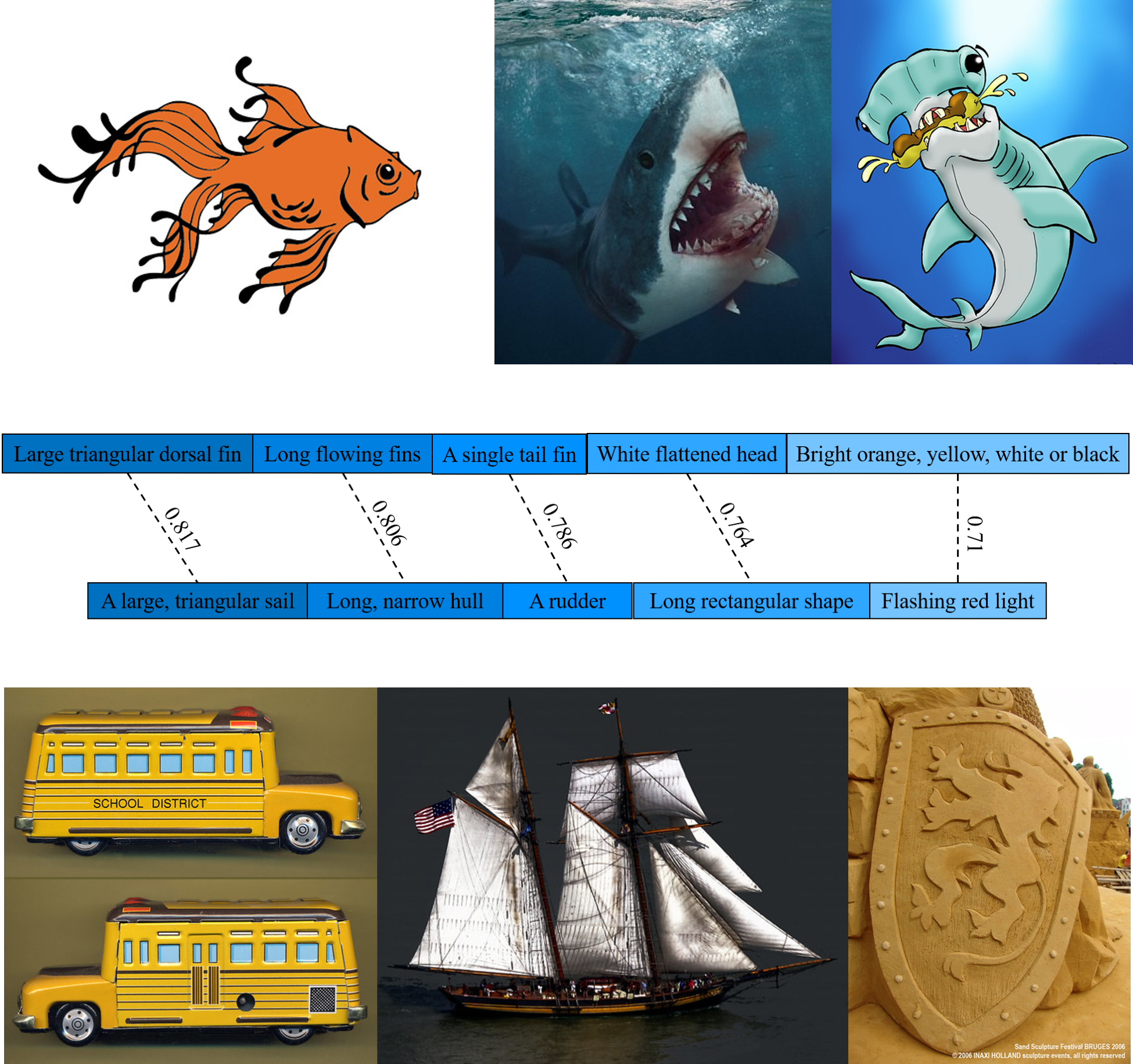}
	\vspace{1mm}
	\caption{{\bf An example of two dissimilar tasks:} {\bf Task 1} contains the classes \emph{goldfish}, \emph{great white shark}, \emph{hammerhead} and {\bf Task 2} contains the classes \emph{school bus}, \emph{schooner}, and \emph{shield}.}
	\label{fig:sim1_figs}
	\vspace{-3mm}
\end{subfigure}
    \caption{\small {\bf Inter-task attribute based task similarity calculation}:
    For each attribute of the new task, the most similar attributes in the old tasks are found, and the corresponding cosine similarity values are computed (some of these have been shown above). The mean of all such max similarities of attributes in the new task, gives the overall task similarity.}
    \label{fig:sim_figs}
\end{figure*}

%% file: CVPR2024_supp/similarity_results.tex
\section{Attribute similarity - A more closer look}
\label{sec:similarity_exp}
We provide a closer look into our class attribute-based task similarity calculation process with the help of some representative examples from the ImageNet-R~\cite{iccv2021imagenetr} dataset, shown in~\ref{fig:sim_figs}. 
The attribute matching mechanism has been described in-detail in {\bf Sec. 4.3} of the main paper.
For this demonstration, we consider three tasks, with each task containing three classes.

The first example~\ref{fig:sim2_figs} contains two similar tasks with Task 1 containing classes - \emph{lorikeet}, \emph{hummingbird}, \emph{toucan} and Task 2 containing classes - \emph{hermit crab}, \emph{flamingo}, \emph{american egret}. Each attribute in the new task is matched with the most similar attribute from the old tasks. The top matching attributes with the highest similarity scores are shown in~\ref{fig:sim2_figs}. The final similarity scores for these two tasks using these attribute similarity scores are obtained to be $0.86$.
The second example~\ref{fig:sim1_figs} contains two relatively dissimilar tasks with Task 1 containing classes - \emph{goldfish}, \emph{great white shark}, \emph{hammerhead} and Task 2 containing classes - \emph{school bus}, \emph{schooner}, \emph{shield}. Each attribute in the new task is matched with the most similar attribute from the old tasks. The top matching attributes (with the highest similarity scores) are shown in~\ref{fig:sim1_figs}. The final similarity scores for these two tasks using these attribute similarity scores are obtained to be $0.70$.
Hence it is evident that our task-similarity technique captures the inter-task similarity well, resulting in a high similarity score for more similar tasks, and a low similarity score for dissimilar tasks.